%% file: ijcai25.tex
\documentclass{article}
\usepackage{ijcai25}

\usepackage{times}
\usepackage{soul}
\usepackage{url}
\usepackage[hidelinks]{hyperref}
\usepackage[utf8]{inputenc}
\usepackage[small]{caption}
\usepackage{graphicx}
\usepackage{amsmath}
\usepackage{amsthm}
\usepackage{booktabs}
\usepackage{algorithm}
\usepackage[switch]{lineno}
\usepackage{mathtools}
\usepackage{amsthm, amssymb}
\usepackage{array}
\usepackage[caption=false,font=normalsize,labelfont=sf,textfont=sf]{subfig}
\usepackage{textcomp}
\usepackage{stfloats}
\usepackage{url}
\usepackage{verbatim}
\usepackage[figuresright]{rotating}
\usepackage{algorithmicx}
\usepackage{algpseudocode} 
\usepackage{multirow}
\usepackage[table,xcdraw]{xcolor}
\usepackage{color}
\usepackage{caption}

\definecolor{mistyrose}{rgb}{1.0, 0.89, 0.88}

\title{Efficient Point Clouds Upsampling via Flow Matching}

\author{
Zhi-Song Liu$^1$
\and
Chenhang He$^2$\thanks{Corresponding Author}
\and
Lei Li$^3$\\
\affiliations
$^1$Lappeenranta-Lahti University of Technology LUT\\
$^2$The Hong Kong Polytechnic University\\
$^3$Technical University of Munich\\
\emails
zhisong.liu@lut.fi,
chenhang.he@polyu.edu.hk,
lilei.cg@gmail.com\\
}

\begin{document}

\maketitle

\begin{abstract}
Diffusion models are a powerful framework for tackling ill-posed problems, with recent advancements extending their use to point cloud upsampling. Despite their potential, existing diffusion models struggle with inefficiencies as they map Gaussian noise to real point clouds, overlooking the geometric information inherent in sparse point clouds. To address these inefficiencies, we propose PUFM, a flow matching approach to directly map sparse point clouds to their high-fidelity dense counterparts. Our method first employs midpoint interpolation to sparse point clouds, resolving the density mismatch between sparse and dense point clouds. Since point clouds are unordered representations, we introduce a pre-alignment method based on Earth Mover's Distance (EMD) optimization to ensure coherent interpolation between sparse and dense point clouds, which enables a more stable learning path in flow matching. Experiments on synthetic datasets demonstrate that our method delivers superior upsampling quality but with fewer sampling steps. Further experiments on ScanNet and KITTI also show that our approach generalizes well on RGB-D point clouds and LiDAR point clouds, making it more practical for real-world applications.
\end{abstract}

\input{Section/01_Introduction}
\input{Section/02_Related_Works}
\input{Section/03_Approach}
\input{Section/04_Experiments}
\input{Section/05_Conclusion}


\bibliographystyle{named}
\bibliography{ijcai25}

\end{document}

%% file: Section/01_Introduction.tex
\section{Introduction}
\label{Introduction}

Point Cloud Upsampling~\cite{punet} has been investigated in recent years as a solution to increase the resolution of the point cloud and enhance the geometric details. Higher-resolution point clouds can substantially benefit a range of downstream tasks, including mesh reconstruction~\cite{p2m}, 3D scene perception~\cite{car,robot} and rendering~\cite{gaussians}.

 As a fundamentally ill-posed problem, point cloud upsampling has traditionally been approached through conventional methods~\cite{opt_1,opt_2}, which often rely on computationally expensive optimization techniques and additional priors, such as surface normals or extra point features. However, these methods tend to struggle in more challenging scenarios, particularly when only sparse points are available, making it difficult to infer the underlying topology. Deep learning approaches~\cite{mpu,grad-pu,pudm,pugan,repkpu} have demonstrated superior performance in point cloud upsampling. By training deep neural networks to encode point cloud geometry into informative deep features, these methods are able to more effectively reconstruct high-resolution point clouds.  Among these, PUNet~\cite{punet} is a pioneering work to learn multi-level features per point and reconstruct the underlying surface based on the contextual information of point set.

\begin{figure}[t]
	\centering
		\centerline{\includegraphics[width=\columnwidth]{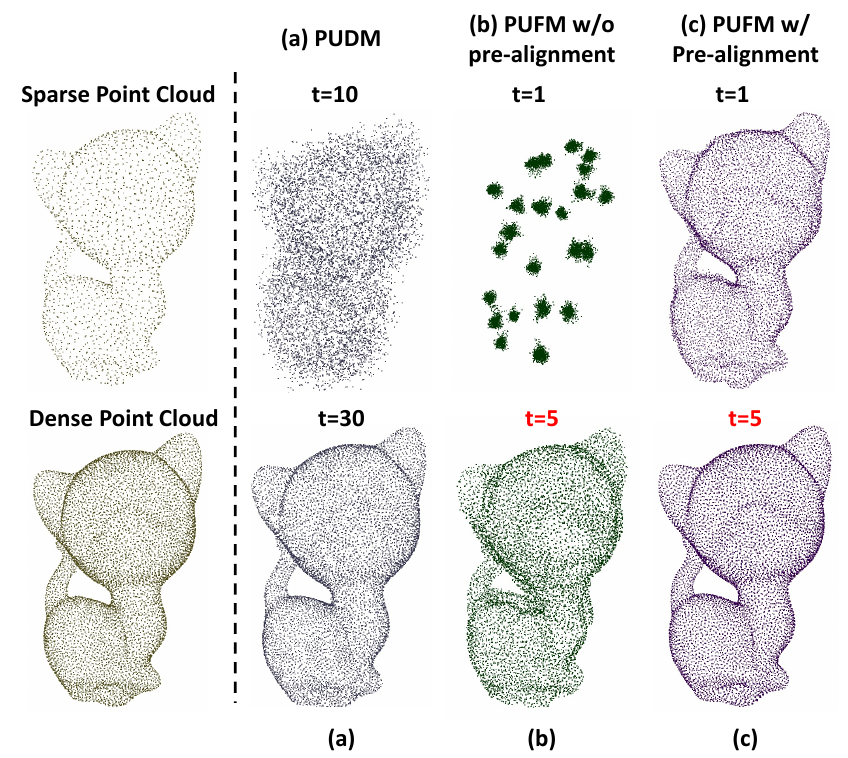}}
		\caption{\textbf{Convergence comparison among different distribution mapping paths.} The diffusion model PUDM demonstrates slow convergence since it starts from noise distribution. Our proposed PUFM learns flow matching from sparse to dense point clouds. Pre-alignment is applied to minimize the learning ambiguity at the early stage, resulting in a more efficient upsampling process.}
		\label{fig:demo}
\end{figure}

Recent advances in diffusion models~\cite{ddpm} for ill-posed problems have motivated their application to point cloud upsampling. PUDM method~\cite{pudm} proposed a conditional probability diffusion model that treats point cloud upsampling as a noise-to-data generative process, conditioned on the sparse point clouds. This approach has demonstrated high-quality dense prediction for both uniform and nonuniform point clouds. However, we argue that this noise-to-data process is inefficient for point cloud upsampling, as the sparse point clouds themselves already encode the valuable prior information about the target  structure. As illustrated in Figure~\textcolor{red}{\ref{fig:demo}a}, PUDM exhibits relatively slow convergence in the reconstruction of dense point clouds, suggesting that the reliance on noise generation may be unnecessarily complex for the task at hand.

In this work, we propose \textbf{Point cloud Upsampling via Flow Matching (PUFM)}, an efficient method that directly learns the optimal transport between sparse and dense point cloud distributions, significantly reducing both the learning complexity and sampling cost compared to diffusion models. To address the cardinality difference between the two distributions, we first apply midpoint interpolation~\cite{grad-pu} to densify the point clouds. However, we observe that the flow matching model encounters ambiguous learning during  early stages, as indicated by the collapsed ball-shaped clusters in Figure~\textcolor{red}{\ref{fig:demo}b}. This issue arises from the inherent unordered nature of point clouds, which leads to multiple bijective matching paths between the sparse and dense point sets. Due to this inherent ambiguity, the flow matching model tends to learn a suboptimal ``average path" during the early steps of training, which slows its convergence. To mitigate this, we propose pre-aligning the sparse point clouds to their corresponding ground-truth by optimizing their Earth Mover's Distance (EMD). This pre-alignment step helps to reduce the ambiguity in the flow matching process by providing a clearer correspondence between the sparse and dense distributions. The learning trajectory in flow matching becomes much smoother and therefore results in faster convergence during training. Importantly, the pre-alignment is not required during inference. As demonstrated in Figure~\textcolor{red}{\ref{fig:demo}c}, the optimized flow matching model is able to reconstruct high-fidelity dense point clouds with only a few sampling steps, achieving both high accuracy and efficiency.

To evaluate the effectiveness of our approach, we apply PUFM to datasets such as PUGAN~\cite{pugan} and PU1K~\cite{pu-gcn}, where it outperforms existing methods, achieving state-of-the-art performance. Furthermore, we extend our method to real-world datasets, including Scannet~\cite{scannet} and KITTI~\cite{kitti}. The qualitative results highlight our method's ability to reconstruct extremely sparse instances, demonstrating its superior generalization and robustness in diverse scenarios.

%% file: Section/02_Related_Works.tex
\section{Related Works}
\label{related_work}
\subsection{Point Clouds Analysis}
Different from image data, point clouds contain irregular structures and are invariant to permutation. PointNet~\cite{pointnet} is the pioneering work that applies shared MLPs to extract point-wise features, and then uses maxpooling for the extraction of permutation-invariant features. PointNet++~\cite{pointnet++} further improves it by proposing set abstraction to extract multi-level features. Similar works can be also found in~\cite{rethinking,pointnext,dgcnn,ldgcnn,kpconv}. For instance, PVCNN~\cite{pvcn} modifies the PointNet++ to handle vocalized point cloud representation, a low rank matrix approximation algorithm is proposed to estimate normals for underlying 3D surface reconstruction. Recently, attention~\cite{attention} has also been widely investigated for point cloud processing. \cite{pct,pct2,fpct,ppt,pointcept,point,pointr,ptv2,ptv3} have achieved impressive performance for point cloud recognition, denoising and enhancement. 

In this work, we build our flow matching model on top of PointNet++ as it encodes multi-level contextual information and produces point-wise representation.

\subsection{Point Clouds Upsampling}
Point cloud upsampling has been extensively studied in recent years, with several approaches proposed to enhance point cloud resolution and improve geometric details. Early learning-based methods, such as PUNet~\cite{punet}, paved the way for using deep learning in point cloud upsampling. MPU~\cite{mpu} introduced a progressive upsampling technique using patch-based processing, while PUGCN~\cite{pu-gcn} employed graph convolution networks for learning local point information via NodeShuffle. PUBP~\cite{dualbp} introduced a back-projection network for iterative down- and up-sampling refinement, and PUGAN~\cite{pugan} leveraged adversarial networks to supervise the generation distribution, achieving more evenly distributed dense points. Further developments include self-supervised learning methods~\cite{sapcu,self} for exploiting local-scale geometric recurrence, as well as methods like PUGeo~\cite{pugeo} and NePs~\cite{np} that project point clouds to 2D for continuous upsampling via convolution. Other works~\cite{octree,grad} focus on combining upsampling and denoising with joint supervision to better integrate point features. Despite these advances, most methods still rely on Chamfer distance, which fails to capture detailed underlying 3D structures in point clouds.

In this paper, we propose a flow matching approach for point cloud upsampling that directly models the transport between sparse and dense point cloud distributions. Unlike previous methods relying on Chamfer distance or displacement-based approaches~\cite{mpu,dispu,repkpu}, our method efficiently learns the mapping between sparse and dense point clouds, which optimize the $l_2$ loss in an iterative inverse path. Our method is also related to recent work in score-based denoising models~\cite{pudm,tp} and normalizing flows~\cite{puflow}, but unlike these approaches, which generate dense point clouds from noise, our method directly generates them from sparse data. By incorporating pre-alignment, we significantly reduce permutation ambiguity, enabling faster and more efficient inference.

%% file: Section/03_Approach.tex
\section{Approach}
\label{approach}

\begin{figure*}[t]
	\centering
		\centerline{\includegraphics[width=\textwidth]{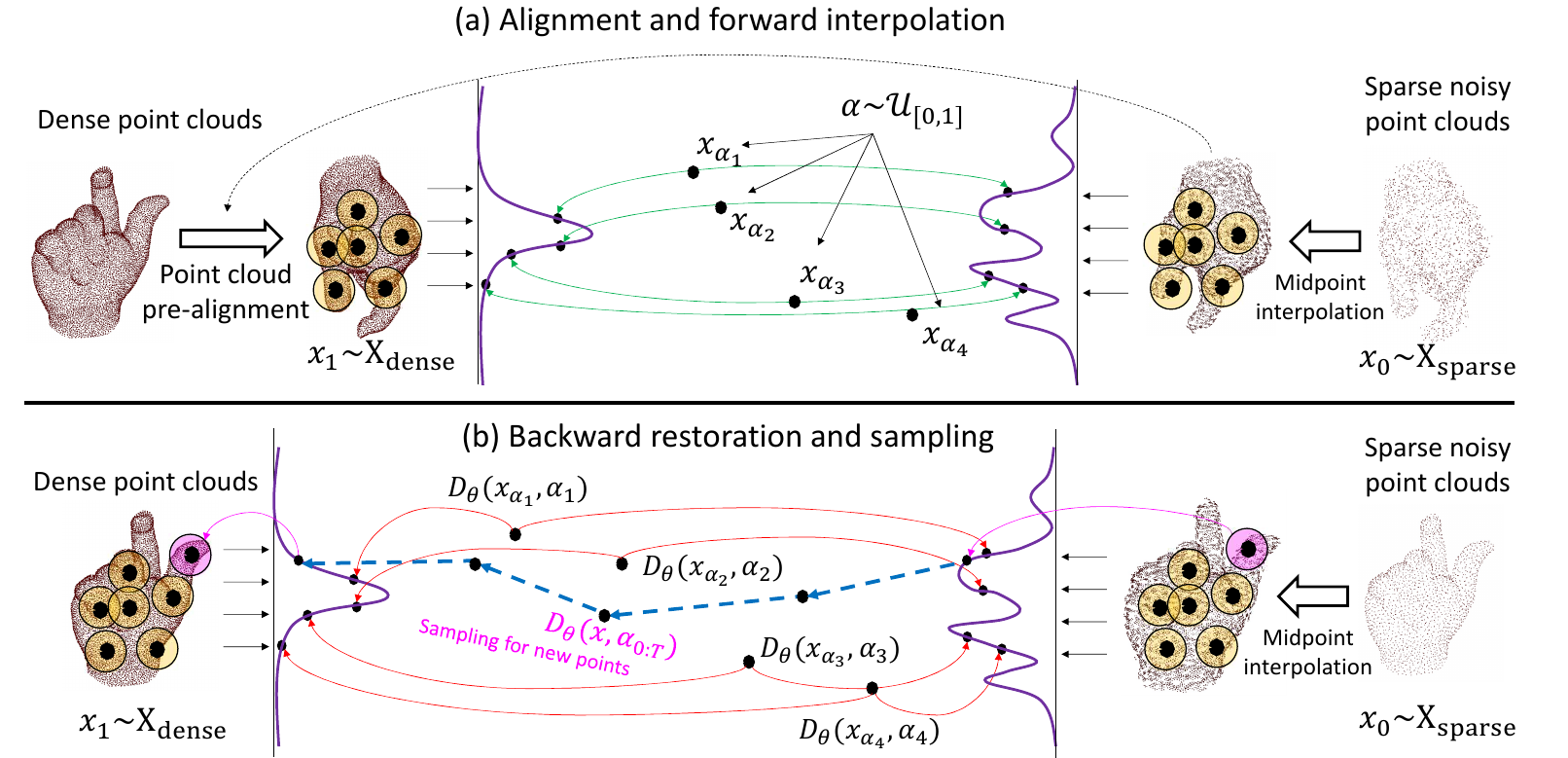}}
		\caption{\small{\textbf{Illustration of PUFM.} It processes point clouds as patches (yellow circles), and learns $D_\theta$ to map the distribution from sparse to dense patches. It contains forward alignment and interpolation (a) and backward restoration and sampling (b). In (a), the model first pre-aligns the sparse (initially upsampled by midpoint interpolation) and dense point clouds and then randomly picks ``noisy'' points as $x_{\alpha_i}$ with the time step $\alpha_i$. In (b), the model optimizes the distribution flow to transform arbitrary sparse data into dense data effectively. During the sampling process, the model directly learns to densify the sparse points without further alignment. Note: the rotated sparse point clouds with 180 degrees illustrate the effect of misalignment between sparse and dense point clouds.
		}}
		\label{fig:network}
\end{figure*}

\subsection{Preliminary}
Given the target data $x\sim p(x)$, we aim to build a flow matching generative model~\cite{flow} to learn the revere time-dependent process to obtain noise $\epsilon\sim N(0, I)$ as,

\begin{small}
\begin{equation}
x_{t} = \alpha_t x + \sigma_t \epsilon, \ \ \text{where}\ t\in [0, 1]
\label{eq:fm_1}
\end{equation}
\end{small}

\noindent where $\alpha_t$ is an increasing function of t and $\sigma_t$ is a decreasing function of t. We expect that the model finds a marginal probability distribution $p_t(x)$ that satisfies $p_0(x)\equiv e(x)$ and $p_1(x)\approx p(x)$. We have the following Ordinary Differential Equation (ODE) to describe the velocity field:

\begin{small}
\begin{equation}
\frac{dp_t(x)}{dt} + \nabla\cdot \left(p_t(x)\nu_\theta(x, t) \right) = 0
\label{eq:fm_2}
\end{equation}
\end{small}

\noindent where $\nu_\theta(x, t)$ can be learned by minimizing the loss,

\begin{small}
\begin{equation}
\mathcal{L}(\theta)=\mathbb{E}_{t\sim\mathcal{U}[0,1]}\left[||(x-\epsilon)-\nu_\theta(x_t, t)||^2\right]
\label{eq:fm_3}
\end{equation}
\end{small}

\noindent With learned $\nu_\theta(x, t)$, we can use the Monte Carlo method to approximate the integration. The inverse sampling of $p_1(x)$ is generated by integrating the ODE from $t=0$ to $t=1$ by solving the discrete ODE by Euler method:

\begin{small}
\begin{equation}
x_{t+1}=x_t+(\delta_{t+1}-\delta_t)\nu_\theta(x_t, t)
\label{eq:fm_4}
\end{equation}
\end{small}

\noindent where $\delta$ is the discrete timestep.

\subsection{Point cloud Upsampling via Flow Matching}
We propose to leverage flow matching for point cloud upsampling. One related work is PUDM~\cite{pudm}, which learns a straightforward noise-to-data model, and then directly optimizes the $l_2$ distance between the true and estimated point clouds. This approach has two limitations: 1) \textit{sparse point clouds already contain partial views of the dense points,  which encodes the underlying structure of the dense point clouds. Learning from noise is computationally costly and inefficient}; and 2) \textit{Point clouds are unordered and have an irregular format, making it challenging for the model to pair points for learning tractable interpolation.} To resolve these issues, we propose a flow matching model PUFM for point cloud upsampling, which directly learns the optimal transport from sparse to dense distribution. To learn the flow matching model, we first densify the sparse point clouds and prealign them with the ground truth based on the Earth Mover's Distance (EMD)~\cite{emd}. The overall framework is depicted in Figure~\textcolor{red}{\ref{fig:network}}. 
\vspace{2mm}

\noindent\textbf{Point cloud flow matching.}
Given the sparse point cloud $x_0\sim \mathcal X_{\text{sparse}}\in \mathbb{R}^{M\times3}$ and dense point clouds $x_1\sim \mathcal X_{\text{dense}}\in \mathbb{R}^{N\times3}$. Our goal is to learn a flow matching model by finding the optimal transport between $\mathcal X_{\text{sparse}}$ and $\mathcal X_{\text{dense}}$. Since the sparse point cloud is a partial view of dense point clouds but shares the same underlying 3D structure, we initially densify the sparse point cloud via a midpoint interpolation~\cite{grad-pu} method. Given the sparse point clouds $x_0$, its densified version $\Tilde{x}_0=\text{mid}(x_0, \eta)$ can be defined as:
\begin{small}
\begin{equation}
\Tilde{x}_0= \frac{1}{2}\cdot[\text{R}_{\gamma}(x_0) + \text{FPS}(x_0, \gamma)] + \eta n, \ \ \text{where}\ n\sim\mathcal{N}(0, I)
\label{eq:mid}
\end{equation}
\end{small}

\noindent where $\text{R}_{\gamma}(\cdot)$ means repeating points by $\gamma$ times and FPS denotes the Furthest Point Sampling~\cite{mpu}. Without losing generalization, we also add a small portion of Gaussian noise with noise level $\eta$ to the sparse point cloud to simulate real-world noisy point clouds. The objective of learning a point cloud upsampling model can then be interpreted as learning the velocity field $\nu$ from two distributions. We parameterize the velocity field by a network $\nu_\theta(x_t, t)$ with learnable parameter $\theta$, and the loss term is defined as:

\begin{equation}
\mathcal{L}(\theta)=\operatorname*{min}_\theta \operatorname*{\mathbb{E}}_{t,\Tilde{x}_0,x_1} \Vert \nu_\theta(x_t, t) - [x_1-\Tilde{x}_0] \Vert^2,
\label{eq:train_loss}
\end{equation}
where $t\in[0,1]$ and $x_t$ are the timestep and the intermediate interpolant of the flow matching respectively.
\begin{figure*}[t]
	\centering
		\centerline{\includegraphics[width=\textwidth]{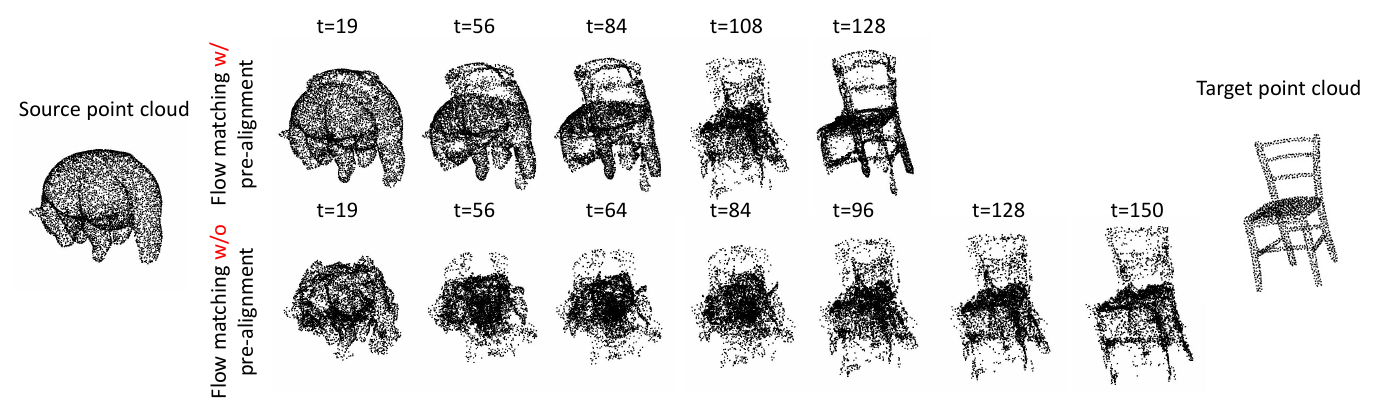}}\vspace{-2mm}
		\caption{\small{\textbf{A toy example on flow matching for point cloud transformation.} Without pre-alignment, the model converges slower as it first transforms the source point clouds to a set of dispersed clusters and then gradually matches to the target point clouds. In contrast, the pre-alignment exhibits more efficient and consistent transformation.
		}}
		\label{fig:toy}
\end{figure*}

\noindent\textbf{Point cloud pre-alignment.}
In this work, we consider the upsamling follows a straight path \cite{lipman2022flow} between sparse and dense distributions, \emph{i.e.}:
\begin{equation}
    x_t = (1-t) \Tilde{x}_0 + t x_1
\end{equation}
However, since the sparse point clouds are randomly sampled from their dense counterparts with additional noise, both $\Tilde{x}_0$ and $x_1$ have an unordered format, which leads to a misalignment between the sparse and dense point clouds. The residual signal $x_1-\Tilde{x}_0$ between two distributions varies significantly, leading to ambiguous learning by $\nu_\theta(x_t,t)$. To minimize Equation ~\textcolor{red}{\ref{eq:train_loss}}, the model finds it easier to spread the points uniformly across space. This causes a set of diffused ball-shaped clusters since the model will first solve these irregularities between these unpaired points by dispersing the points to a more tractable distribution, rather than approaching to the target distribution.

To illustrate this phenomenon, we conduct a toy experiment to learn a flow matching model that maps from the ``bear" distribution to the ``chair" distribution. As illustrated in Figure~\textcolor{red}{\ref{fig:toy}}, the model without pre-alignment will disperse the points at the early stage, and the output is also blurry and noisy. In contrast, the model with the pre-alignment demonstrates a consistent transformation from sparse to dense point clouds. It is worth noting that \textbf{the pre-alignment is applied only in the training phase}. Our model demonstrates higher efficiency as a result.

To this end, we propose to use the Earth Mover’s Distance (EMD) optimization~\cite{emd} to find an optimum assignment $\phi^*=\phi(\Tilde{x}_0, x_1)$ that minimizes the average distance for each point in the point cloud $\mathbf{x}_{1}$ to its nearest neighbor in $\mathbf{x}_{0}$:
\begin{equation}
\phi^*=\operatorname*{argmin}_\phi \Phi \sum_{i=1}^N \Vert x_1^{\phi(i)}-\Tilde{x}_0^i \Vert_2,
\label{eq:interpolate}
\end{equation}
where $\Phi={{1,...,N}\rightarrow {1,...,N}}$ is the set of possible bijective assignments between points in $x_1$ and $\Tilde{x}_0$. During the training stage, we first apply $\phi^*$ to permute the dense point clouds, aligning them with the sparse point clouds. This alignment ensures that the model learns from a consistent pairing of sparse and dense point clouds. Next, we sample the interpolant $x_t$ from the paired data.  As shown in Figure~\textcolor{red}{\ref{fig:demo}}, our approach demonstrates faster convergence when pre-alignment is applied, and high-fidelity dense point clouds can be generated in just a few sampling steps. In contrast, when pre-alignment is not applied, the model encounters early-stage collapse of the point clouds, hindering effective learning.

\noindent \textbf{Training and sampling}
Based on our previous discussion, we now have the complete training scheme and sampling process. Furthermore, the distribution of $t$ used during training has a clear impact on performance. We choose $t=1-cos(s\pi/2),s\sim\mathcal{U}[0,1]$ to encourage the model to be certain of the direction to move at the very early steps of the procedure. A comparison of different sampling schemes can be found in the supplementary section. The complete training and sampling process is shown in Algorithm \textcolor{red}{1} and \textcolor{red}{2}.\vspace{2mm}

\begin{algorithm}
\caption{Training}
\begin{algorithmic}
\Require $x_0 \sim \mathcal X_{sparse}$, $x_1 \sim \mathcal X_{dense}$, $t \sim \mathcal{U}{[0,1]}$
\State $n\sim \mathcal{N}(0, I)$
\State $\Tilde{x}_0 = \text{mid}(x_0, \eta)$ \Comment{midpoint interpolation Eq~\textcolor{red}{\ref{eq:mid}}}
\State $\phi^*=\phi(\Tilde{x}_0, x_1)$ \Comment{EMD alignment Eq \textcolor{red}{8}}
\State $x_t \gets (1 - t)\Tilde{x}_0 + t x_1^{\phi(i)}$
\State $l \gets ||\nu_\theta(x_t, t) - (x_1^{\phi(i)} - \Tilde{x}_0)||^2$
\end{algorithmic}
\end{algorithm}
\begin{algorithm}
\caption{Sampling}
\begin{algorithmic}
\Require $x_0 \sim \mathcal X_{sparse}$, $\delta$
\State $x_t = \text{mid}(x_0, \eta)$
\For{$t = 0$ \textbf{to} $1$}

    \State $x_{t+1} = (1- \frac{\delta}{t})x_t + \frac{\delta}{t} \nu_\theta(x_t, t)$
\EndFor
\end{algorithmic}
\end{algorithm}

\noindent \textbf{Model architecture}
We follow previous works~\cite{ddpm_pc,pudm,p2p} and use a model architecture based on PointNet++~\cite{pointnet++}. It is a UNet structure that aggregates set abstraction (SA) and feature propagation (FP) blocks together to learn multiscale point features. We add the vector attention blocks~\cite{pct,pct2} at the last layer of SA blocks to encourage global feature mapping. The network is conditioned on the timestep $t$ using sinusoidal positional embeddings. Both SA and FP blocks are included with conditional MLP layers that can incorporate the time embedding for feature extraction. The details of the model structure are shown in the supplementary.

%% file: Section/04_Experiments.tex
\section{Experiments}
\label{experiment}

\begin{figure*}[t]
	\centering
		\centerline{\includegraphics[width=\textwidth]{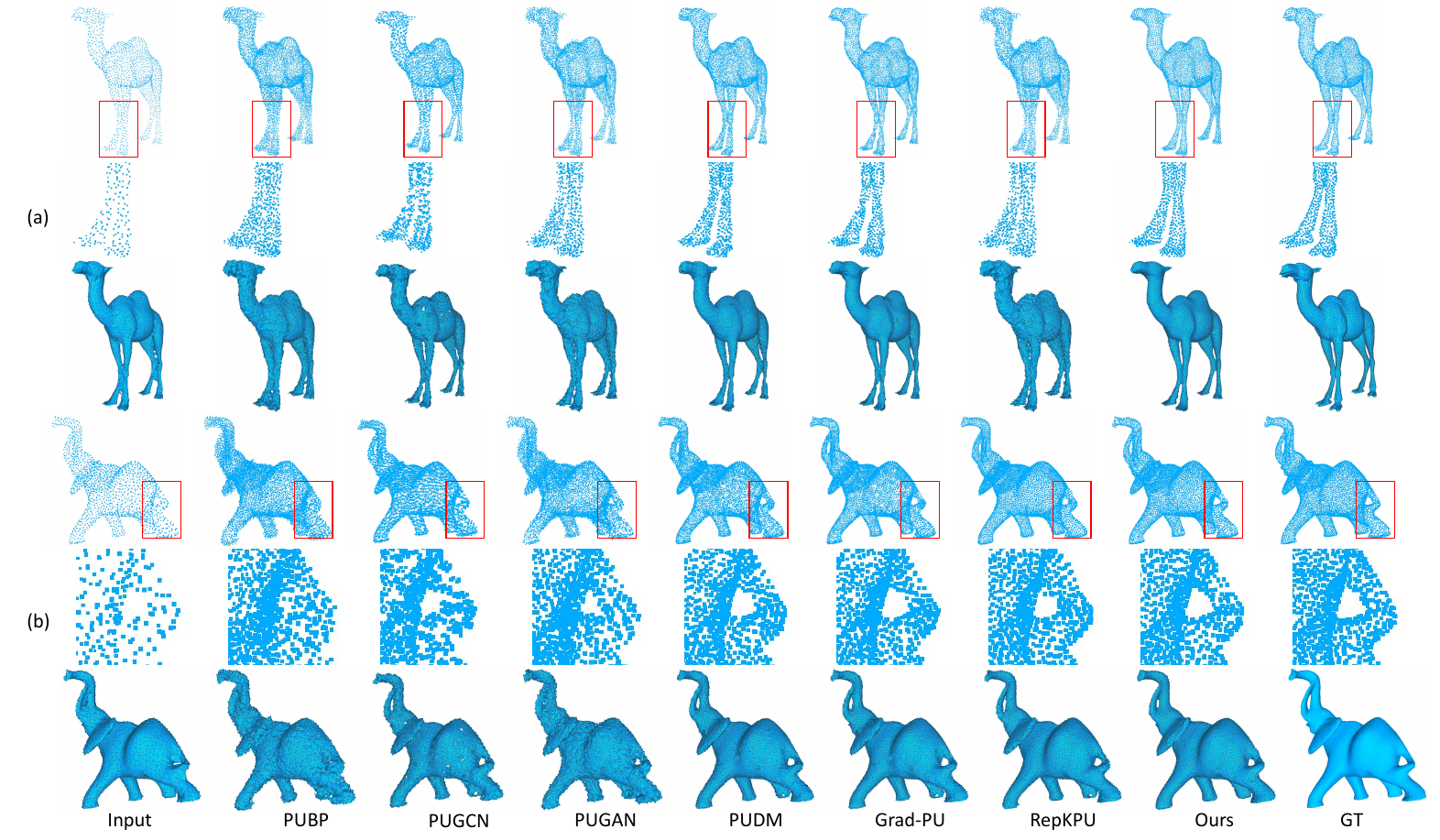}}
		\caption{\small{\textbf{Visual comparison of different methods on 4$\times$ upsampling.} We zoom in on the red box regions to highlight the point cloud upsampling differences. We also show the mesh reconstruction, and ours produces evenly distributed point clouds and smooth meshes. 
		}}
		\label{fig:sota}
\end{figure*}

\begin{table}[t]
\centering
\renewcommand\arraystretch{1.3}
\resizebox{\linewidth}{!}{
\begin{tabular}{llccc|ccc}
\toprule
\multicolumn{2}{c}{Upscaling factors} & \multicolumn{3}{c|}{4$\times$} & \multicolumn{3}{c}{16$\times$} \\
\cmidrule(lr){3-5} \cmidrule(lr){6-8}
Dataset & Method & CD & HD & P2F & CD & HD & P2F \\
\midrule
\multirow{7}{*}{PUGAN} 
 & PUBP   & 1.649 & 1.476 & 5.997 & 0.982 & 2.071 & 7.496 \\
 & PUGCN  & 2.774 & 3.831 & 9.508 & 1.102 & 1.785 & 7.125 \\
 & RepKPU & 1.067 & 1.139 & 1.974 & 0.384 & 1.245 & 2.151 \\
 & PUGAN  & 1.541 & 1.391 & 5.420 & 0.869 & 1.746 & 6.757 \\
 & PUDM   & 1.221 & 1.174 & 3.132 & 0.533 & 1.185 & 3.589 \\
 & Grad-PU & 1.132 & 1.186 & 1.957 & 0.415 & 1.142 & 2.185 \\
 & \cellcolor{mistyrose}{Ours}   & \cellcolor{mistyrose}{\textbf{1.049}} & \cellcolor{mistyrose}{\textbf{0.876}} & \cellcolor{mistyrose}{\textbf{1.864}} & \cellcolor{mistyrose}{\textbf{0.353}} & \cellcolor{mistyrose}{\textbf{0.844}} & \cellcolor{mistyrose}{\textbf{2.103}} \\
\midrule
\multirow{7}{*}{PU1K} 
 & PUBP   & 0.694 & 0.593 & 2.206 & 0.343 & 0.712 & 2.958 \\
 & PUGCN  & 1.241 & 1.504 & 5.115 & 2.393 & 4.214 & 9.410 \\
 & RepKPU & 0.566 & 0.619 & 1.464 & 0.290 & 0.592 & \textbf{1.821} \\
 & PUGAN  & 0.682 & 0.632 & 2.792 & 0.361 & 0.773 & 3.538 \\
 & PUDM   & 0.706 & 0.605 & 2.891 & 0.421 & 0.602 & 3.370 \\
 & Grad-PU & 0.626 & 0.583 & 1.510 & 0.316 & \textbf{0.552} & 1.890 \\
 & \cellcolor{mistyrose}{Ours}   & \cellcolor{mistyrose}{\textbf{0.545}} & \cellcolor{mistyrose}{\textbf{0.556}} & \cellcolor{mistyrose}{\textbf{1.770}} & \cellcolor{mistyrose}{\textbf{0.220}} & \cellcolor{mistyrose}{0.578} & \cellcolor{mistyrose}{1.983} \\
\bottomrule
\end{tabular}%
}
\caption{\textbf{Overall comparison on point cloud upsampling.} We report CD ($10^{-4}$), HD ($10^{-3}$) and P2F ($10^{-3}$) on different methods. Ours outperforms baselines under different upsampling factors on different datasets.}
\label{tab:sota}
\end{table}

\subsection{Experimental Details}
\noindent \textbf{Dataset.} We use two public datasets for training and evaluation: PUGAN~\cite{pugan} and PU1K~\cite{pu-gcn}. We follow the same procedure in ~\cite{mpu} to extract paired sparse and dense point clouds. Given the 3D meshes, we first use Poisson disk sampling to generate uniform patches as ground truth, each patch contains 1024 points. Then we randomly sample 256 points from each patch to obtain the sparse input data. For testing, we obtain 27 and 127 point clouds from PUGAN and PU1K, respectively. Code will be released soon.

\noindent \textbf{Evaluation metrics.} We use the Chamfer Distance (CD), Hausdorff Distance (HD) and Point-to-Surface (P2F) as the evaluations in our experiments.

\subsection{Comparison with State-of-the-arts}
We quantitatively evaluate our method with several point cloud upsampling approaches, including PUBP~\cite{dualbp}, PUGCN~\cite{pu-gcn}, PUGAN~\cite{pugan}, PUDM~\cite{pudm}, Grad-PU~\cite{grad-pu} and RepKPU~\cite{repkpu}. For fair comparison, we use the training parameters and model weights provided in the publicly available code bases for previous works. For the evaluation, we use randomly sampled sparse point clouds as input to compare different methods in $4\times$ and $16\times$ upsampling. In Table~\textcolor{red}{\ref{tab:sota}}, we show the overall comparison to other state-of-the-art methods. We can see that ours achieves the best CD, HD and P2F scores, which indicates that ours delivers high-fidelity point clouds better than diffusion based models and other learning based approaches.

For visualization, Figure~\textcolor{red}{\ref{fig:sota}} shows the $4\times$ upsampling results using different methods. To better visualize the point distributions, we highlight the parts in the red boxes, and we can see that ours can produce evenly distributed dense points, while others still produce noise or outlier points that do not follow the underlying surface. To further illustrate the upsampling quality, we apply the PCA normal estimation (neighborhood number = 15)~\cite{normal} and ball pivot approach (depth=9)~\cite{ball} to all upsampling point clouds for surface reconstruction. We can see that ours can produce smoother surfaces without showing any holes. For example, other approaches produce a hole in the leg of the elephant and noisy surfaces of the camel, while ours produces smooth results.

\begin{figure}[t]
	\centering
		\centerline{\includegraphics[width=\columnwidth]{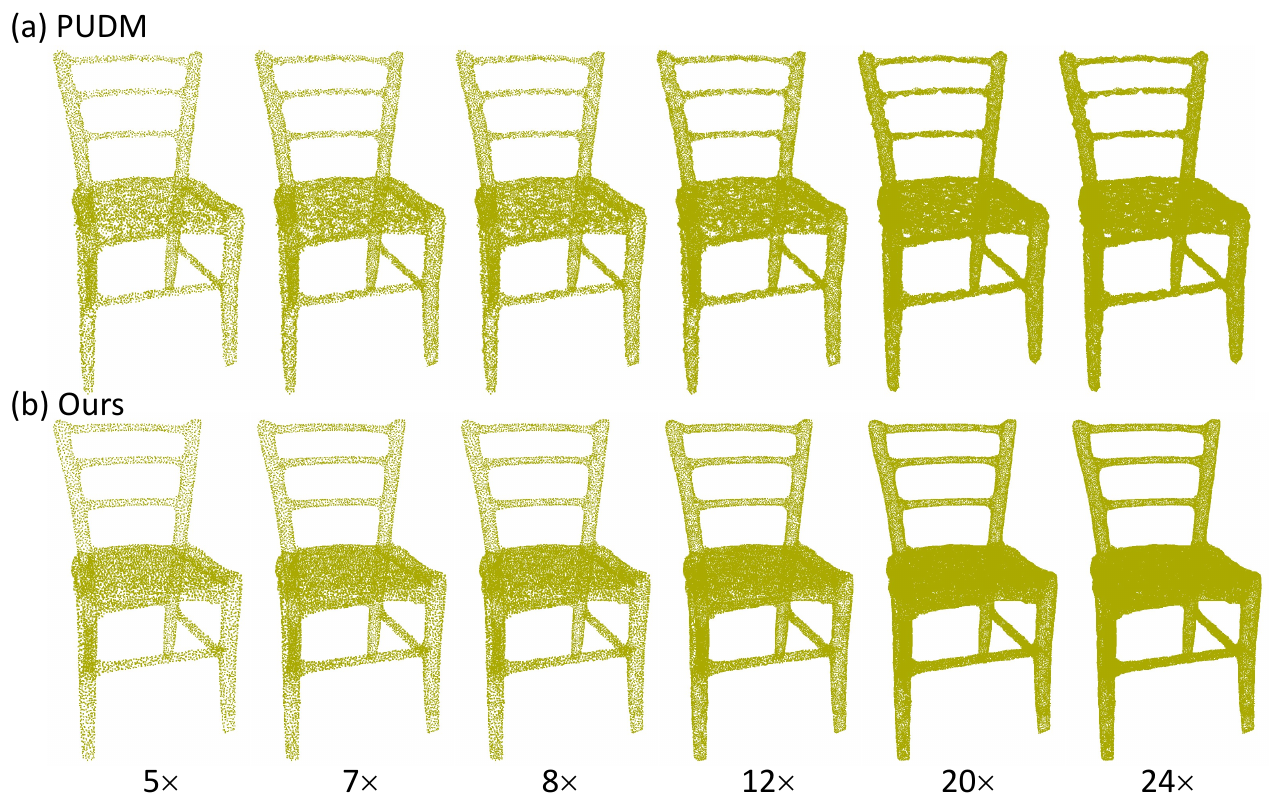}}
		\caption{\small{\textbf{Visulization of arbitrary point cloud upsampling.} 
		}}
        \vspace{-3mm}
		\label{fig:arbitrary}
\end{figure}

\begin{table}[t]
\centering
\renewcommand\arraystretch{1.}
\resizebox{\linewidth}{!}{
\begin{tabular}{c|cccccc}
\toprule
\multirow{2}{*}{Method} & \multicolumn{2}{c}{Grad-PU} & \multicolumn{2}{c}{PUDM} & \multicolumn{2}{c}{\cellcolor{mistyrose}{Ours}} \\
 & CD & HD & CD & HD & \cellcolor{mistyrose}{CD} & \cellcolor{mistyrose}{HD} \\ \midrule
5$\times$ & 0.952 & 4.841 & 1.054 & 5.091 & \cellcolor{mistyrose}{\textbf{0.921}} & \cellcolor{mistyrose}{\textbf{4.763}} \\
7$\times$ & 0.752 & 4.523 & 0.850 & 4.537 & \cellcolor{mistyrose}{\textbf{0.708}} & \cellcolor{mistyrose}{\textbf{4.363}} \\
8$\times$ & 0.686 & 3.880 & 0.784 & 4.054 & \cellcolor{mistyrose}{\textbf{0.638}} & \cellcolor{mistyrose}{\textbf{3.947}} \\
12$\times$ & 0.515 & 3.628 & 0.617 & \textbf{3.371} & \cellcolor{mistyrose}{\textbf{0.455}} & \cellcolor{mistyrose}{3.380} \\
20$\times$ & 0.452 & 3.238 & 0.473 & 2.695 & \cellcolor{mistyrose}{\textbf{0.344}} & \cellcolor{mistyrose}{\textbf{2.666}} \\
24$\times$ & 0.412 & 3.108 & 0.444 & 2.691 & \cellcolor{mistyrose}{\textbf{0.314}} & \cellcolor{mistyrose}{\textbf{2.690}} \\
32$\times$ & 0.365 & 2.667 & 0.404 & \textbf{2.322} & \cellcolor{mistyrose}{\textbf{0.273}} & \cellcolor{mistyrose}{2.517} \\ \bottomrule
\end{tabular}%
}
\caption{\textbf{Arbitrary point clouds upsampling on PUGAN.} Ours can outperform baselines in CD and HD across different upsampling factors.} 
\label{tab:arbitrary}
\end{table}

\begin{figure}[t]
	\centering
		\centerline{\includegraphics[width=\columnwidth]{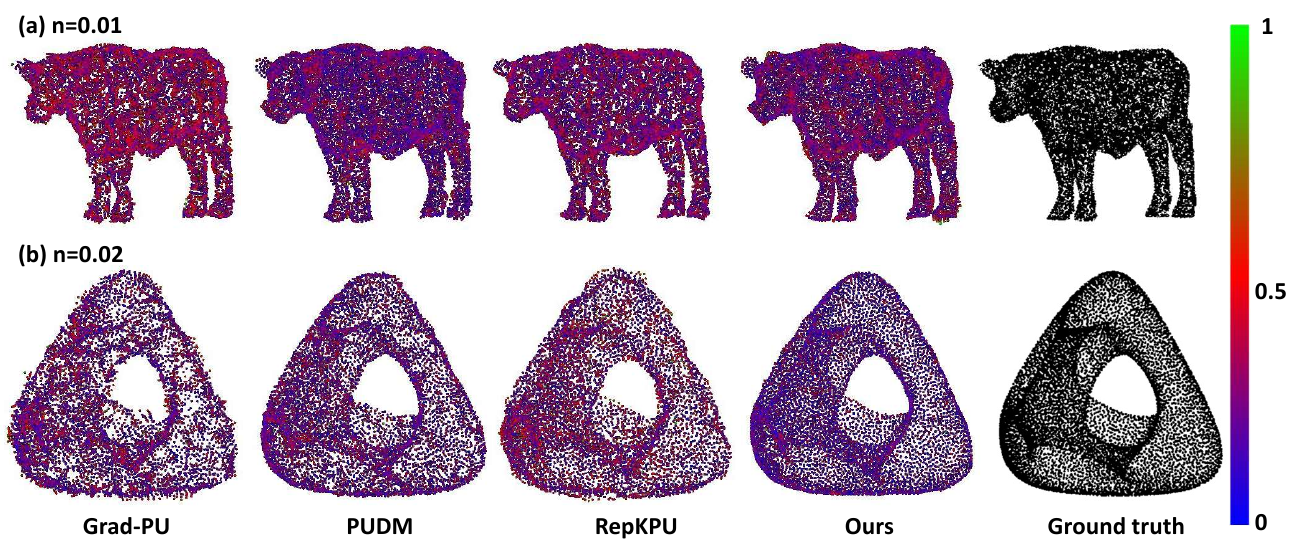}}
		\caption{\small{\textbf{Visulization of noisy point cloud upsampling.} We calculate the P2F distance as the color feature to the point cloud for visualization. Ours are visually better than baselines, without showing holes in (a) and noisy outliers in (b).
		}}\vspace{-3mm}
		\label{fig:noise}
\end{figure}

\begin{figure}[t]
	\centering
		\centerline{\includegraphics[width=\columnwidth]{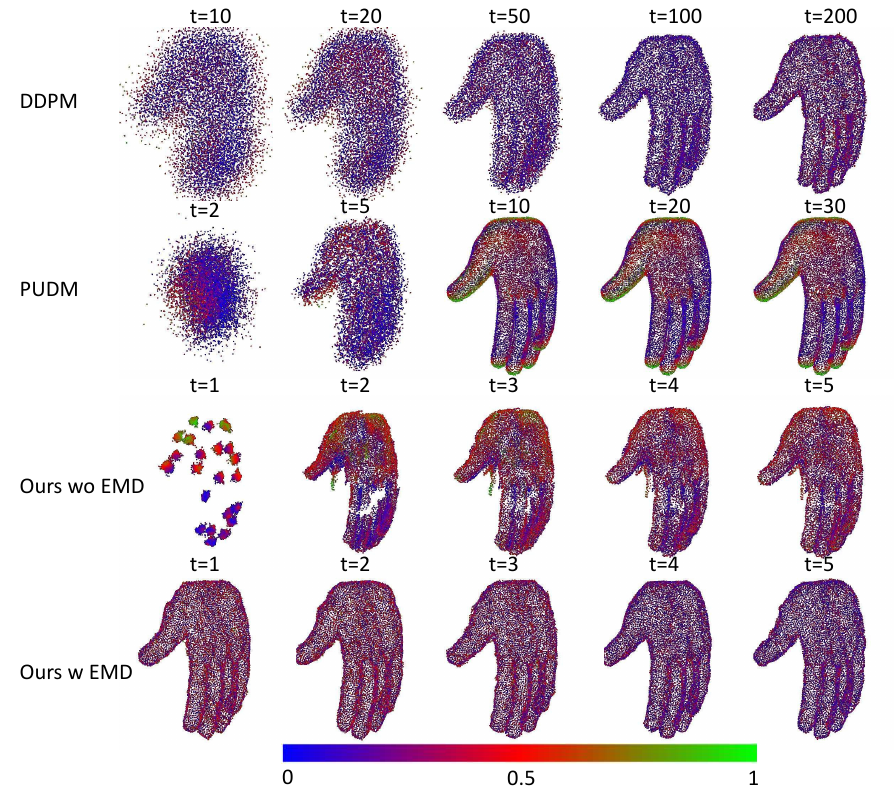}}
		\caption{\small{\textbf{Sample visualization at different time steps.} We compare with DDPM, PUDM, and ours at different sampling steps. Ours quickly learns the optimal upsampling point clouds, while others take longer sampling steps ($>10$). 
		}}
		\label{fig:time_step}
\end{figure}

\begin{table}[t]
\centering
\renewcommand\arraystretch{1.3}
\resizebox{\linewidth}{!}{
\begin{tabular}{c|ccccccc}
\hline
\multirow{2}{*}{Dataset} & \multicolumn{3}{c}{PUGAN} & \multicolumn{3}{c}{PU1K} & \multirow{2}{*}{\begin{tabular}[c]{@{}c@{}}Running time (s)\\ sampling steps\end{tabular}} \\
 & CD & HD & P2F & CD & HD & P2F & \\ \hline
DDPM & 1.239 & 1.558 & 3.207 & 0.618 & 0.739 & 2.616 & 7.83/100 \\
InDI & 1.542 & 1.529 & 6.033 & 0.737 & 0.730 & 3.903 & 0.82/10 \\
PUDM & 1.221 & 1.174 & 3.132 & 0.706 & 0.605 & 2.891 & 1.83/30 \\
Ours wo EMD & 2.817 & 2.442 & 3.446 & 1.188 & 0.845 & 3.045 & 0.71/5 \\
\cellcolor{mistyrose}{Ours w EMD} & \cellcolor{mistyrose}{\textbf{1.049}} & \cellcolor{mistyrose}{0.876} & \cellcolor{mistyrose}{\textbf{1.864}} & \cellcolor{mistyrose}{\textbf{0.545}} & \cellcolor{mistyrose}{\textbf{0.556}} & \cellcolor{mistyrose}{\textbf{1.770}} & \cellcolor{mistyrose}{\textbf{0.71/5}} \\ \hline
\end{tabular}%
}
\caption{\textbf{Ablation study on different inversion models.} We report the results on PUGAN and PU1K using different inversion models. The running time is calculated based on the optimal sampling steps reported in the corresponding papers.}
\label{tab:dm}\vspace{-2mm}
\end{table}

\begin{table}[t]
\centering
\renewcommand\arraystretch{1.3}
\resizebox{\linewidth}{!}{
\begin{tabular}{cccc|ccc}
\hline
Noise level & \multicolumn{3}{c|}{$\eta=$0.01} & \multicolumn{3}{c}{$\eta=$0.02} \\ \cline{2-7} 
Method & CD & HD & P2F & CD & HD & P2F \\ \cline{2-7} 
PUDM & 1.706 & \textbf{1.803} & 6.025 & 3.100 & 3.622 & 1.150 \\
RepKPU & 1.696 & 2.189 & 6.989 & 3.974 & 4.827 & 1.615 \\
Grad-PU & 1.795 & 1.965 & 6.591 & 3.588 & 4.248 & 1.310 \\
\cellcolor{mistyrose}{Ours} & \cellcolor{mistyrose}{\textbf{1.496}} & \cellcolor{mistyrose}{1.879} & \cellcolor{mistyrose}{\textbf{5.887}} & \cellcolor{mistyrose}{\textbf{2.404}} & \cellcolor{mistyrose}{\textbf{2.794}} & \cellcolor{mistyrose}{\textbf{1.055}} \\ \hline
\end{tabular}%
}
\caption{\textbf{4$\times$ point cloud upsampling on noisy PUGAN.} Our performs better than others at different noise levels, indicating its robustness against noise.}
\label{tab:noise}\vspace{-2mm}
\end{table}

\subsection{Arbitrary upsampling and denoising}
Our proposed method can produce multi-scale point cloud upsampling as others~\cite{grad-pu,pudm} by repeatedly applying 4$\times$ upsampling and FPS downsampling. To show the versatility of ours, we compare ours with Grad-PU and PUDM on 5$\times$ to 32$\times$ upsampling and report the results in Table~\textcolor{red}{\ref{tab:arbitrary}}. We can see that ours steadily outperforms the other two methods in CD, HD, and P2F. We also show \textit{Chair} example from PUGAN in Figure~\textcolor{red}{\ref{fig:arbitrary}}. We observe that PUDM produces holes on the surface of the chair, while ours can produce evenly distributed points without deviating from the underlying surface.

\subsection{Ablation Studies}
As discussed in the previous section, ours is closely related to the diffusion model and inversion by direct iteration. To demonstrate the efficiency of our method, we compare with DDPM~\cite{ddpm}, PUDM~\cite{pudm} and InDI~\cite{indi}. Note that DDPM and InDI were originally for image processing, and we modified them to achieve point-cloud upsampling. Table~\textcolor{red}{\ref{tab:dm}} shows that ours reduces the CD distance by approximately 0.1$\sim$0.2 points and HD distance by approximately 0.1$\sim$0.2  points. We also show the optimal sampling steps and the corresponding running time based on corresponding papers. We can see that ours uses the least sampling steps and achieves fast computation. This is due to the PC-to-PC flow matching which can reduce the restoration iterations. To demonstrate the impact of the proposed point cloud pre-alignment, we can see from row 4 and 5 that using proposed EMD pre-alignment is important for the flow matching model to start with reasonable point pairs for restoration. Without using the point cloud pre-alignment, the upsampling quality in all metrics drops significantly. This supports our discussion in Section 3.2 that the pre-alignment approximate the local region for point cloud upsampling, such that PUFM can model the distribution close to the ground truth.

Figure~\textcolor{red}{\ref{fig:time_step}} shows the intermediate samples in different time steps. We represent the point clouds by encoding the P2F distances as RGB colors. We can see that ours can quickly converge to the ground truth point clouds with fewer errors (indicated by the blue color). We also observe that without using point cloud pre-alignment, the model first collapses into several clusters and then gradually reaches the ground truth data. Qualitative results, runtime, and experiments with other model setups are provided in the supplementary materials.

\subsection{Robustness and downstream applications}
From Table~\textcolor{red}{\ref{tab:sota}}, the most competitive methods are RepKPU, Grad-PU and PUDM. However, ours is more resilient to noise. To demonstrate the robustness of our method, we report the results in Table~\textcolor{red}{\ref{tab:noise}}. Given the PUGAN dataset, we randomly add Gaussian noise with different levels ($\eta$) to apply 4$\times$ upsampling. We can see that ours consistently performs better at different noise levels, with approximately 0.3$\sim$1.1 drops in terms of CD, HD and P2F. We are also interested in the upsampling performance on real-world data. We test ours and others on ScanNet~\cite{scannet} and KITTI~\cite{kitti} datasets. Without the ground truth data, we visualize the upsampled point clouds in Figure~\textcolor{red}{\ref{fig:real}}. We can see that ours performs better than PUDM in (a) of the bicyclist without showing gaps in the human body. In (b), ours can produce more evenly distributed dense points, revealing the indoor 3D layouts. We believe that directly using sparse point clouds as the prior for upsampling, ours remains consistent with real-world structures. The model does not need to hallucinate details entirely from Gaussian noise. More comparisons can be found in the supplementary.

%% file: Section/05_Conclusion.tex
\begin{figure}[h]
	\centering
     \vspace{-3mm}\centerline{\includegraphics[width=\columnwidth]{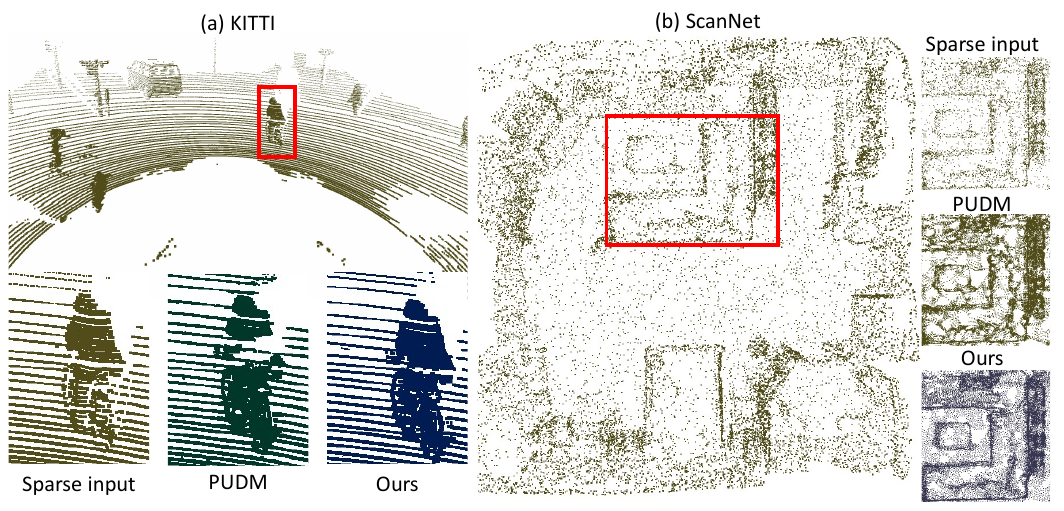}}	\caption{\small{\textbf{Visulization of real point cloud upsampling.} We use two examples from ScanNet and KITTI to apply 4$\times$ upsampling. Compared to PUDM, ours does not show inconsistent patterns on the bicyclist or noisy point distribution in the living room.
		}}
        \vspace{-3mm}
		\label{fig:real}
\end{figure}

\section{Conclusion}
\label{conclusion}
In this paper, we introduce Point cloud Upsampling via Flow Matching (PUFM), an efficient framework that learns the direct optimal transport between sparse and dense point clouds. Unlike existing approaches that rely on direct $l_2$ optimization, which often struggle with the unordered and irregular nature of point clouds, our method employs the optimization with EMD to pre-align sparse and dense point clouds before applying probability-based interpolation. Comprehensive experiments on both synthetic and real-world datasets demonstrate that PUFM consistently outperforms state-of-the-art methods in terms of accuracy and robustness. Additionally, PUFM achieves high efficiency compared to diffusion-based approaches and demonstrates strong resilience to noise and scalability across different input densities.